\documentclass{article}

\usepackage{times}
\usepackage{epsfig}
\usepackage{graphicx}
\usepackage{amsmath}
\usepackage{amssymb}

\begin{document}
\title{Resolution Enhancement of Range Images\\via Color-Image Segmentation}

\author{Arnav Bhavsar\ \\PhD, Indian Institute of Technology Madras, India}

\date{}

\maketitle

\begin{abstract}
We propose a method for super-resolution of range image. Our approach leverages the interpretation of LR image as sparse samples on the HR grid. Based on this interpretation, we demonstrate that our recently reported approach which reconstructs dense range images from sparse range data by exploiting a registered colour image, can be applied for the task of resolution enhancement of range images. Our method only uses a single colour image in addition to the range observation in the super-resolution process. Using the proposed approach, we demonstrate super-resolution results for large factors (e.g. 4) with good localization accuracy.   
\end{abstract}


% \fi
%\IEEEpeerreviewmaketitle

\section{Introduction}
Within the last decade range cameras and scanners are established as an important dominant acquisition modality in computer vision and multimedia community. Applications where range cameras are successfully employed include digital heritage \cite{heritage}, industrial inspection, filming and 3DTV \cite{3dtv}, gaming \cite{kinect} etc. Range cameras operate on a variety of technologies involving laser scanning \cite{yang,parrot}, time-of-flight imaging \cite{arnav,lidar} and structured or coded lighting \cite{middlebury} which determine aspects such as accuracy, resolution, speed of acquisition and cost. For instance, there exists laser-based and structured-light-based scanners provide accurate and high resolution (HR) range scans, but the acquisition is typically slow and requires a good amount of manual effort. On the other hand, the low-cost laser scanners, time-of-flight scanners and light-coding range scanners (e.g. Kinect \cite{kinect}) acquire range images much faster and with little manual intervention, but are limited in accuracy and resolution. Such a trade-off motivates the development of computational approaches to enhance the resolution and accuracy, so as to make range scanning more affordable, less time-consuming.

The low spatial resolution is a common issue with low-cost range scanning. Often, the acquired range image resolution is of the order of 100 to 200 pixels (in one direction) \cite{arnav}. As compared to digital cameras (which typically span megapixels), these acquisition grids are too small to capture sufficient scene detail, edges etc. Hence, the resolution needs to enhanced by large factors (e.g. 4, 6 etc. ). Naive approaches of enhancing the resolution by image interpolation methods results in heavy loss of accuracy. As a result, recent years has seen the rise of various sophisticated approaches \cite{yang,seg,example} to enhance the resolution of range images which involve preserving edges as well as maintaining good accuracy.

A key to enhance the spatial resolution of range images by large factors, is to utilize a high-resolution registered colour image of the same scene, as done in these approaches. The colour image can be easily captured using any off-the-shelf digital camera. \footnote{The registration can be achieved via one-time pre-calibration of the digital and range cameras or using any of the well established landmark-based registration approaches. Similar to the above referred approaches \cite{yang,seg}, we too assume that the range and colour images are pre-registered.} The use of an HR colour image is motivated by an observation that commonly, the prominent depth discontinuities (such as those between different objects in a scene) coincide with colour discontinuities \cite{stereoseg}. Thus, information about the presence/absence of colour discontinuities and local similarity of colour information, which is derived from the HR colour image helps in localizing range discontinuities and estimating dense range information on the HR grid. Such a cue from the colour image has also been used in many stereo disparity estimation approaches. 

In this work, we notice that the problem of range super-resolution (SR) can, in fact, be interpreted as that of reconstructing a range image from sparsely captured data. In this context, we recently reported an approach from reconstruction of dense range maps from sparsely captured range data \cite{arnav}, which uses a similar cue from the registered colour image via its segmentation. The advantages of this approach include a relatively simple local estimation approach, good accuracy, and computational efficiency.  Here, we demonstrate that this approach would also hold for the task of range super-resolution. 

%However, based on the discussion in \cite{arnav}, and based on our own experimentation with the approach, we realize that the sparsity for the SR task by large factors (e.g. by 8) is a little too much so as to be handled by an direct application of the method in \cite{arnav}. In this work, we propose an approach built upon this approach to enable super-resolution of range images by large factors.

The paper is organized as follows. In the next subsection we discuss some related work. Section 2 covers our methodology, which includes a brief description of the basic approach in \cite{arnav}.
%followed by our proposed extensions for he SR task. 
We then provide some results in section 3, and conclude in section 4.


%\hfill January 11, 2007

\subsection{Related work}
As mentioned earlier, the idea of using a colour image has been explored for range super-resolution in various ways. For example, the work in \cite{henrik} interpolates the range image and, by exploiting the assumption that depth discontinuities coincide with color edges, improves estimation at discontinuities. Similar improvements are shown in MRF-based energy minimization approaches that also uses a HR optical image \cite{huhle,mrf}. However, these approaches are not known to perform for large super-resolution factors. 

\begin{figure*}[!t]
\centering
\begin{tabular}{c c}
\includegraphics[width=40pt]{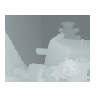} & 
\includegraphics[width=160pt]{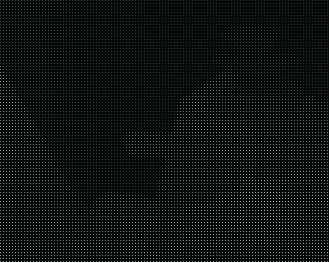} 
\end{tabular}
\caption{\label{fig:sparserep} (a) LR range image and, (b) its corresponding sparse HR interpretation. The HR grid is 4 $\times$ the LR grid in each direction.}
\end{figure*}

The authors in \cite{yang} propose an application of bilateral filtering which exploits constraints from the HR colour image. This work demonstrates ability to achieve good quality super-resolution by large factors. However, the bilateral filter, which is based on the HR colour image, has to be defined at each pixel. This makes the approach computationally very demanding. 
%For instance, our Matlab implementation of the method takes about 2 hours for a HR images with sizes of about 500 $\times$ 500 pixels. 
Our approach which works on image segments rather than pixels, and is based on computing local costs over segments, is a lot more efficient.
%takes about 2 minutes on the same image sizes, and also provides very low errors.

Another recent approach which also employs a colour image is reported in \cite{seg}. This method also uses the segmentation of the colour image. However, beyond this, it significantly differs from our method. The range labels are computed using a modified version of bilinear interpolation which is directed by the colour segmentation. On the other hand, ours approach computes range labels based on minimization of local costs with regularization constraints.

More recently, an example based range super-resolution approach is also reported \cite{example}, which also performs well at large resolution factors but requires a separate range dataset from which the training examples are derived. In contrast, our approach, like the above mentioned methods, only requires a registered colour image.

\section{Resolution enhancement as reconstruction of sparse range data}

As mentioned earlier, our approach is based upon the idea of treating the super-resolution problem as that of range reconstruction from sparse data \cite{arnav}. This is inspired by the idea that LR range images can be essentially interpreted as down-sampled versions of their HR counterparts, where the down-sampling is obtained by selecting range pixels at regular intervals (depending on the SR factor). That is, the LR image is modeled as a decimation of the HR image. While in most image super-resolution approaches, the down-sampling is modeled by averaging of the HR pixels, the decimation model for down-sampling is not uncommon for range super-resolution \cite{schuon1,lidar}. Thus, observing an LR image can be interpreted as observing only those samples of the HR image which were selected during the down-sampling process, with spaces between these samples having a zero value on the HR grid, thus resulting in a sparse HR image. An example of the sparse interpretation of an LR range image is shown in Fig. \ref{fig:sparserep}, where for a resolution enhancement with a factor of 4, close to 93$\%$ of the data missing. Such large magnitudes of missing data are also considered in \cite{arnav}. Hence, the SR task is essentially that of reconstructing such a sparse HR image. 

%However, in our experiments, we noticed that the sparse configurations for large super-resolution factors cannot be handled well by the method of \cite{arnav} alone. (For instance, for a SR factor of 4, about 98$\%$ for the pixels on the HR grid have a zero value). Hence, we consider additional improvements more specific to the task of super-resolution. In this section, before discussing these improvements, we will first briefly recap the method of \cite{arnav}.

\subsection{Proposed method}

As mentioned we begin by representing the low resolution range image on the high-resolution image grid by uniformly spacing the pixels in the low resolution range on the HR grid. The associated colour image is of the same size of this HR image grid.

The method starts with segmentation of the registered colour image. The colour segmentation is carried out using the well-known mean-shift algorithm (MSA) \cite{ms}. While a detailed description of the MSA is beyond the scope of this paper, it is worthwhile to state that the MSA uses two kernels defined on the colour dimension and the spatial dimension to determine the coarseness of the final colour segmentation. The greater these bandwidths, the coarser the segmentation and bigger the segments. These two parameters are important in the context of the approach in \cite{arnav}. At the beginning, the method starts with small bandwidths and as a result achieves a largely over-segmented image, where each object is divided into multiple segments.

Following the initial colour segmentation, the approach computes plane-fit for each segment, based on the available range pixels in the segment. However, this is done only if the number of visible range pixels ($n_a$) that segment exceeds a threshold $n_{pl}$ ($n_a \geq n_{pl}$). The plane-fitting is carried out using the RANSAC method \cite{ransac}. Given the fitted-plane over a segment, a local cost is defined for assigning a range label $z$ for each invisible pixel $p$ in the segment as
\begin{eqnarray}\label{eq:plfitcost}
C_p = |z-z_{pl}| + \lambda_p\sum_{q\in V_p} |z-z_q|
\end{eqnarray}
Here, $z_{pl}$ is the plane-fitted range at $p$ and $V_p$ is the set of \emph{visible} second-order neighbours of $p$ that belong to segment $s$. The first term computes distance from the plane-fitted range. The second term, weighted by $\lambda_p$, enforces similarity between neighbours. The label estimated for the pixel $p$ is simply the one which minimizes $C_p$.

If for a segment $0 < n_a < n_{pl}$, then plane-fitting may not be robust enough. For such segments, a median range $z_m$ over the $n_a$ pixels is computed. In addition, the medians of the visible pixels over the adjacent segments are also computed. Based on this, a cost is defined as
\begin{eqnarray}\label{eq:mediancost}
C_s = |z-z_m| + w_a\sum_{z_{m_a}\in m_a}|z-z_{m_a}|
\end{eqnarray}
where $m_a$ is the set of the medians $z_{m_a}$ of the visible pixels of $a\in A_s$. In equation \ref{eq:mediancost}, the second term enforces similarity over neighbouring segments. 
The weight $w_a$ is defined for a pair of adjacent segments $s$ and $a$ as follows
\begin{eqnarray}\label{eq:contextwt}
w_a = \frac{\lambda_m}{(|\overline{r_s}-\overline{r_a}| + |\overline{g_s}-\overline{g_a}| +  |\overline{b_s}-\overline{b_a}|)}
\end{eqnarray}
Here, $\overline{r_i}$, $\overline{g_i}$, $\overline{b_i}$ are the mean RGB intensities for segment $i$. This contextual weighting strengthens the smoothness between similarly coloured segments but weakens it for segments with large colour differences. The label minimizing $C_s$ is assigned to all the missing pixels in the segment $s$. Recall that this assignment is only for segments for which $0 < n_a < n_{pl}$. These are typically small segments for whom a constant surface assumption is quite valid.

As one would have observed, the above process only labels the pixels in the segments for which $n_a > 0$. However, there will be many segments with $n_a = 0$, when large number of pixels are missing. Just one pass of the above process, with a constant colour segmentation will leave many pixels unlabeled. %However, iterating on only the cost computation, keeping the segmentation 
%constant, will not help either since the segments with na = 0 will
%never be considered while computing Cp or Cs.
To resolve this issue, the mean-shift segmentation, in each iteration, is performed with different (slightly larger) kernel bandwidths, so that as iterations progress, the segmentation produces somewhat larger segments. This process essentially subsumes many of the currently labeled and unlabeled segments into common larger segments. Thus, in subsequent iterations, many of the currently unlabeled segments are no longer isolated, thus making their pixels eligible for labeling. 
%As mentioned in Section 2.1, the parameters hs and hI control
%the mode resolution in the mean-shift method. Higher the bandwidths,
%coarser the mode resolution and larger the segments. Thus,
%increasing hs and hI in each iteration facilitates segment expansion.
%For simplicity, we only increase the intensity bandwidth hI over
%iterations.
%Note that although the segments expand, they expand non-uniformly
%such that similar and smaller segments have more probability
%of being grouped. Moreover, we vary hI such that the
%segment expansion is gradual. Thus, typically, smaller segments
%that are spatially close are gradually merged; the plane-fittinglabeling
%is performed more often thus capturing the gradual range
%variations. 
Since similar segments are more likely to be grouped, the demarcation between prominent range discontinuities that corresponds to largely different segments, is still maintained.
Thus, the segment expansion process allows the cost computation modules to label all the pixels over iterations while maintaining prominent range discontinuities. A complete labeling is typically achieved in 4-5 iterations.

\section{Results}

\begin{figure*}[!t]
\centering
\begin{tabular}{c c c c c c}
\hspace{-1cm}
\includegraphics[width=160pt]{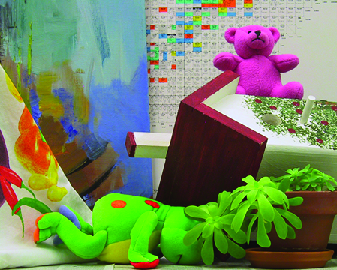} & 
\includegraphics[width=160pt]{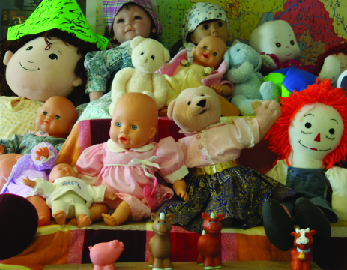} &
\includegraphics[width=160pt]{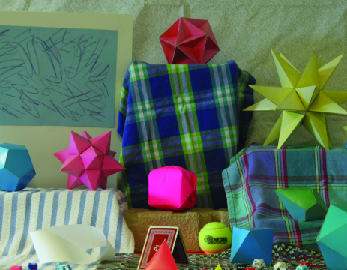}\\\hspace{-1cm}(a)&(b)&(c)\\
\hspace{-1cm}
\includegraphics[width=40pt]{teddy_lr_4_1.png} & 
\includegraphics[width=40pt]{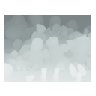} &
\includegraphics[width=40pt]{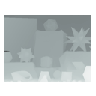}\\\hspace{-1cm}(d)&(e)&(f)
\end{tabular}
\caption{\label{fig:colourimg} (a,b,c) Colour images from the Middlebury dataset used in our experiments. (d,e,f) Corresponding low resolution range images}
\end{figure*}

\begin{figure*}[!t]
\centering
\begin{tabular}{c c c c c c c c c}
\hspace{-1cm}
\includegraphics[width=160pt]{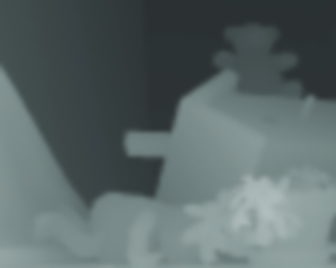} & 
\includegraphics[width=160pt]{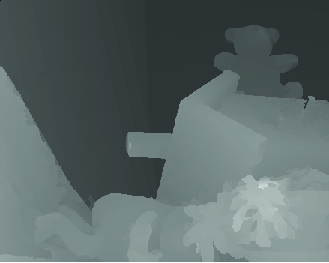} &
\includegraphics[width=160pt]{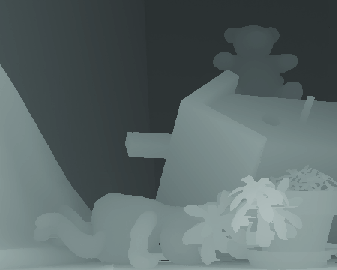}\\\hspace{-1cm}(a)&(b)&(c)\\
\hspace{-1cm}
\includegraphics[width=160pt]{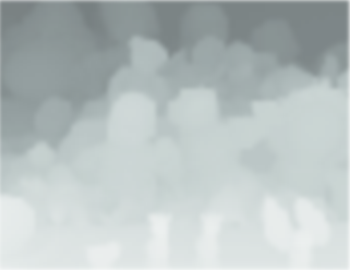} & 
\includegraphics[width=160pt]{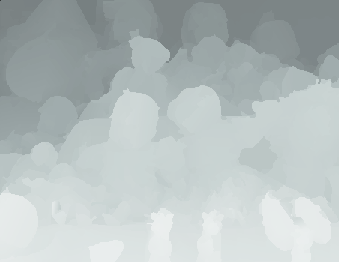} &
\includegraphics[width=160pt]{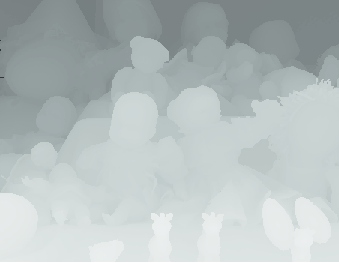}\\\hspace{-1cm}(d)&(e)&(f)\\
\hspace{-1cm}
\includegraphics[width=160pt]{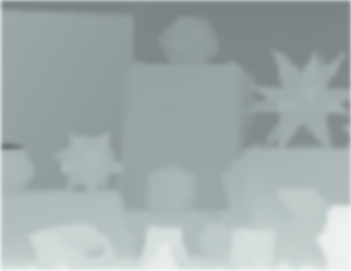} & 
\includegraphics[width=160pt]{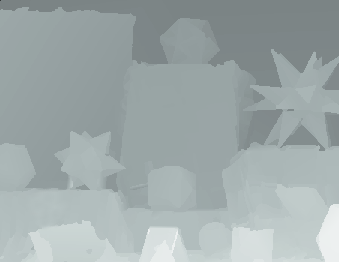} &
\includegraphics[width=160pt]{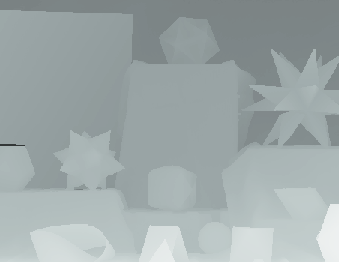}\\\hspace{-1cm}(g)&(h)&(i)
\end{tabular}
\caption{\label{fig:upby4} SR by a factor of 4: (a,d,g) Bicubic interpolation. (b,e,h) Super-resolved range image using our approach. (c,f,i) Ground-truth range.}
\end{figure*}

We now provide some results for our range super-resolution method described in the previous section. We conducted experiments with range images from the Middlebury dataset \cite{middlebury}, which also has colour images associated with the range data. Some examples of the colour images corresponding to the scenes used in our experiments are shown in Figs. \ref{fig:colourimg}(a,b,c). The original resolution of the range images is of the order of 400 $\times$ 400 pixels. We down-sample these images and use the down-sampled versions as LR range images. We then compare our SR results with the original range images (or ground truth). 

%Along with the visual results, we also provide quantitative results i.e. mean absolute error (MAE) and percentage mean absolute error (PMAE). The PMAE is defined as MAE$/(max_r - min_r)$, where $max_r$ and $min_r$ are the maximum and minimum range values in the ground truth. Thus, the PMAE determines the error relative to the extent of the range values.

The results for the case of SR by a factor of 4 (Fig. \ref{fig:upby4}). The low-resolution images are shown in Figs. \ref{fig:colourimg}(d,e,f). One can observe that it is difficult to distinctly perceive the image content in the LR images. Note that for a factor of 4, the bicubic-interpolated image is very blurred at the edges (Figs. \ref{fig:upby4}(a,d,g)). Hence, the object shapes clearly lack proper localization. In comparison, our super-resolution result (Fig. \ref{fig:upby4}(b,e,h)) shows clear improvements over the interpolated image. Such an improvement in the localization is the essence of range super-resolution. Moreover, in addition, one can also notice that high-level of fidelity when the super-resolved range images are compared with the ground-truth (Fig. \ref{fig:upby4}(c,f,i)). As mentioned earlier, the factor of 4 case has a total of 93$\%$ missing data in its sparse interpretation. Inspite of this, our approach is able to maintain localization and fidelity in the eventual result. 
%The accuracy is also justified by the very low MAE and PMAE. Note that over the extent of the range value the percentage error is less than 1$\%$.

%As we had implied earlier, we are mainly interested in super-resolution by large factor. Hence, we next show some results for super-resolution by a factor of 8 (Fig. \ref{fig:upby8}). In this case, the LR image sizes are of the order of 60 $\times$ 60. Clearly, going from this resolution to original resolution is a formidable task. In the sparsity context, the LR pixels are placed on the HR grid with a gap of 8 pixels (in both directions) between two available pixels. This makes about 98$\%$ of data is missing in the original problem of a factor of 8 super-resoluion. Indeed, the interpolated images show heavy blurring and overlap among various objects and hence an obvious lack of clarity (Figs. \ref{fig:upby8}(a,d,g)). On the other hand, our approach still maintains good localization (Figs. \ref{fig:upby8}(b,e,h)).  Even for more complex scenes (e.g. Figs. \ref{fig:upby8}(e,h))), the shapes and edges of most of the scene objects are well-maintained and clearly defined. While the accuracy is understandably lesser than that in the `up-by-4' case, the drop in the accuracy is very little, and our outputs still yeilds very low errors even for such high factors of super-resolution.

Lastly, for completeness, we reassert that our super-resolution approach takes less than a couple of minutes for each of the above cases, on a Xeon 3.2 GHZ CPU with 12 GB RAM, with a Matlab implementation. Thus, we believe that our approach also performs well from the point of view computational efficiency. The efficiency of our method can be attributed to its local nature (i.e. there is no global energy minimization), as well as its segment-based (and not an explicit pixel-based) processing.

\section{Conclusion}
In this work, we investigated the problem of range super-resolution from the point of view of reconstructing dense range maps from sparse data. Based on the realization of LR images as sparse samples at the HR grid, we employed a recently proposed method for sparse range reconstruction, and demonstrated is applicability for the task of range super-resolution. Our approach shows promising results even for large resolution factors such as 4. In future, it would be interesting to gauge the performance of the approach and investigate further improvements under noise and non-exact registration between the range and colour image. % conference papers do not normally have an appendix

%\section*{Acknowledgment}



%\bibliographystyle{IEEEtran}

\end{document}